\documentclass{article}
\usepackage{amsmath,graphicx,mlspconf}
\usepackage{geometry}
\usepackage{amssymb}
\usepackage{xcolor}
\usepackage{comment}
\usepackage{algorithm}
\usepackage{algpseudocode}
\usepackage{url}

\copyrightnotice{978-1-7281-6338-3/21/\$31.00 {\copyright}2021 IEEE}

\toappear{2021 IEEE International Workshop on Machine Learning for Signal Processing, Oct.\ 25--28, 2021, Gold Coast, Australia}

\title{TRANSFER BAYESIAN META-LEARNING VIA WEIGHTED FREE ENERGY MINIMIZATION}
\name{Yunchuan Zhang, Sharu Theresa Jose, Osvaldo Simeone\thanks{This work was supported by the European Research Council (ERC) under the European Union's Horizon 2020 Research and Innovation Programme (grant agreement No. 725732).}}
\address{King's Communications, Learning and Information Processing (KCLIP) Lab,\\ Department of Engineering, King's College London, London WC2R 2LS, UK\\ \{yunchuan.zhang, sharu.jose, osvaldo.simeone\}@kcl.ac.uk}

\def\trm {\mathrm{t}}

\begin{document}

\maketitle

\begin{abstract}
Meta-learning optimizes the hyperparameters of a training procedure, such as  its initialization, kernel, or learning rate, based on data sampled from a number of auxiliary tasks. A key underlying assumption is that the auxiliary tasks -- known as \emph{meta-training tasks} -- share the same  generating distribution as the tasks to be encountered at deployment time -- known as \emph{meta-test tasks}. This may, however, not be the case when the test environment differ from the meta-training conditions.  To address shifts in task generating distribution between meta-training and meta-testing phases, this paper  introduces \emph{weighted free energy minimization} (WFEM) for transfer meta-learning. We instantiate the proposed approach for non-parametric Bayesian regression and classification via Gaussian Processes (GPs). The method is validated on a toy sinusoidal regression problem, as well as on classification using miniImagenet and CUB data sets, through comparison with standard meta-learning of GP priors as implemented by PACOH. 
\end{abstract}
\section{Introduction}
\label{sec:intro}
\textit{Meta-learning }or \textit{learning-to-learn}  aims to extract knowledge from a number of auxiliary tasks so as to speed up learning a new, related task \cite{schmidhuber1987evolutionary, vilalta2002perspective}. For example, consider the problem of training an image classifier for personalized medical diagnosis on a smart phone. By observing data from other individuals, meta-learning can extract knowledge that allows for a fast adaptation on limited data available for a new user of the service. Information across tasks is shared via meta-learning hyperparameters such as an embedding space shared across tasks \cite{vinyals2016matching}, an initialization of a neural network \cite{finn2017model}, or a prior on the weights of a stochastic neural network \cite{amit2018meta}.

An underlying assumption in meta-learning is that the observed auxiliary tasks, known as \textit{meta-training tasks}, and the new, previously unseen \textit{meta-test task} are ``related", in the sense that they belong to the same \textit{task environment}. The task environment defines a distribution over the space of data-generating distributions, and the meta-training and meta-test tasks are assumed to be sampled i.i.d. from the same environment \cite{baxter2000model}. However, this assumption does not hold in many practical scenarios \cite{jose2020transfer}. For instance, in the personalized medical diagnosis example, meta-training data may come from a hospital specializing in patients affected by a specific condition (e.g., cancer patients), while  patients at deployment time may not share the same medical history. 

A meta-learner trained on tasks from a task environment, such as the hospital in the example above, may not perform well on an out-of-distribution (OOD) meta-test task. Recently, reference \cite{jose2020transfer} introduced the problem of \textit{transfer meta-learning}, which accounts for OOD meta-test tasks by modelling the meta-test environment as being distinct from the meta-training environment. In \cite{jose2020transfer}, the authors present PAC Bayes theoretical bounds on the generalization performance of a transfer meta-learner.  

Inspired by the theory developed in \cite{jose2020transfer}, in this work, we introduce an approach for transfer meta-learning termed \textit{weighted free energy minimization (WFEM)} that leverages data from both meta-training and meta-test environments. We specifically focus on non-parametric Bayesian learning via Gaussian processes (GPs), and aim to meta-learn a GP prior to be used on meta-test tasks. WFEM generalizes the information meta-risk minimization (IMRM) for transfer meta-learning introduced by the theory in \cite{jose2020transfer} to non-parametric base-learners, as well as the PACOH-GP based Bayesian meta-learning approach of \cite{rothfuss2020pacoh}. 

Through experiments on real-world and synthetic data sets, in both classification and regression settings, we demonstrate the advantages of transfer meta-learning over conventional learning and meta-learning schemes. Under \textit{meta-environment shift} between the training and testing task environments, we show that the data from the meta-training environment can help improving the predictive performance of transfer meta-learner on the meta-test task as compared to conventional meta-learning schemes such as PACOH-GP. 

\noindent \emph{Related Work:} The problem of meta-learning, when training and testing tasks belong to the same task environment, has been extensively studied both from theoretical perspective \cite{ pentina2014pac, jose2020informationtheoretic} and practical applications \cite{park2020meta}. The difference between the training and testing environments in meta-learning has been accounted for in the recent works of \cite{collins2020taskrobust}, \cite{lee2019learning} that develop meta-learning algorithms robust to meta-environment shift. To the best of our knowledge, the work in \cite{jose2020transfer} is the first to formally introduce the problem of transfer meta-learning, and to obtain theoretical generalization performance guarantees for arbitrary meta-learners.

Bayesian approaches to meta-learning have become increasingly popular in the recent years due to their important advantages in quantifying uncertainty and model selection  \cite{kim2018bayesian,yoon2018bayesian}. In this context, both parametric methods such as Bayesian neural networks and non-parametric methods like GPs have both been successfully applied to real-world applications. The works in \cite{rothfuss2020pacoh,fortuin2019meta}  apply meta-learning to optimize the mean and kernel functions of the GP prior via parametric functions -- a method referred to as PACOH-GP. Our work extends PACOH-GP to transfer meta-learning. The recent paper \cite{rothfuss2021metalearning} presents a modification of  PACOH-GP that operates directly in the functional space.

\section{Problem Setting}
\label{sec:meta}
In this paper, we focus on non-parametric Bayesian learning for regression and classification, whereby the prior is meta-learned  using data from multiple related tasks. In this section, we review the framework introduced in \cite{rothfuss2020pacoh} that defines Bayesian meta-learning as the minimization of a free energy functional. In the next section, we extend this framework to transfer meta-learning by leveraging the theoretical results in \cite{jose2020transfer}. To this end, we first review non-parametric Bayesian learning via Gaussian Processes (GPs), and then we describe the problem of meta-learning the GP prior proposed in \cite{rothfuss2020pacoh}. 
\noindent \emph{Notation:} We use Roman fonts to indicate random variables, functions, and vectors; while the corresponding regular font denotes fixed realizations.
\subsection{Gaussian Processes (GPs)}
\label{ssec:GP}
 We study supervised learning problems. Accordingly, let $\mathcal{X}=\{x_m\}_{m=1}^M$ denote a set of $M$ observed input values in $\mathbb{R}^d$, and let $\mathcal{Y}=\{y_m\}_{m=1}^M$ the corresponding scalar outputs. Each tuple $(x_m,y_m)$ is assumed to be generated i.i.d according to an unknown population distribution $P$. We also denote $\mathcal{D}=\{(x_m,y_m)\}_{m=1}^M$ as the observed training data set. 

We follow a non-parametric Bayesian supervised learning framework, whereby the input-output relationship is modelled by a random function $\trm(x)$ of input $x \in \mathbb{R}^d$ in the presence of observation noise.
Specifically, the observed output $\mathrm{y}_m$ is modelled as a noisy observation of the scalar function $\mathrm{t}(x_m)$ so that, conditioned on $\mathrm{t}(x_m)=t_{m}$, we have the distribution
\begin{equation}
    \mathrm{y}_m\sim p(y_m|t_m) \label{eq:predictor}
\end{equation}
for a fixed conditional distribution $p(y|t)$. As an example, for regression, one may set $p(y|t)=\mathcal{N}(y|t,\sigma^2)$. 

Observations are modelled as conditionally i.i.d. Therefore, letting $\trm(\mathcal{X})$ be the $M\times 1$ vector of outputs of the  random scalar function $\trm(\cdot)$, i.e., $\trm(\mathcal{X})=[\trm(x_1),\hdots,\trm(x_M)]^T$, we have the conditional distribution \begin{equation}
    p(\mathcal{Y}=y|\mathrm{t}(\mathcal{X})=t)=\prod_{m=1}^Mp(y_m|t_m), \label{eq:data_likelihood} \vspace{-0.35cm}
\end{equation}where $t=[t_{1},...,t_{M}]^{T}$.

Adopting a Bayesian  approach, the scalar function $\trm(x)$ is assumed to be random, and is endowed with a GP prior characterized by a mean function $\mu_{\theta}(x)$ and a kernel function $k_{\theta}(x,x')$ \cite{rasmussen2003gaussian}, which we denote as \begin{align}
    \trm(x) \sim \mathcal{GP}(\mu_{\theta}(x),k_{\theta}(x,x')). \label{eq:prior}
\end{align} The GP prior \eqref{eq:prior} is parameterized in terms of a \emph{hyperparameter vector} $\theta$ that determines the mean function $\mu_{\theta}(\cdot)$ and the kernel function $k_{\theta}(\cdot,\cdot)$. Accordingly, the GP defines a prior joint distribution on the output values  $\trm(\mathcal{X})$  as 
\begin{equation}
     p_\theta(\mathrm{t}(\mathcal{X})=t)=\mathcal{N}(t|\mu_{\theta}(\mathcal{X}),K_{\theta}(\mathcal{X})), \label{eq:prior_1}
\end{equation}
where $\mu_{\theta}(\mathcal{X})=[\mu_{\theta}(x_1),...,\mu_{\theta}(x_M)]^T$ is the $M \times 1$ mean vector, and $K_\theta(\mathcal{X})$ represents the $M\times M$ covariance matrix whose $(i,j)$th entry is given as $[K_\theta(\mathcal{X})]_{i,j}=k_\theta(x_i,x_j)$.
Using the GP prior in \eqref{eq:prior_1} and the Gaussian data likelihood $p(y|t)=\mathcal{N}(y|t,\sigma^2)$, the posterior distribution of the random function $\trm(x)$ at a new test input $x$ can be obtained as \cite{rasmussen2003gaussian}
\begin{align}
    p_{\theta}(\mathrm{t}(x)&=t|\mathcal{D})\sim \mathcal{N}(t|\mu(x), s^2(x)), \hspace{0.1cm} \mbox{where}\label{eq:predictive}\\
    \mu(x)&=\mu_\theta(x)+k_{\mathcal{D}}(x)^T(\tilde{K}_\theta(\mathcal{X}))^{-1}(\mathcal{Y}-\mu_{\theta}(\mathcal{X})),\\
     s^2(x)&=k_{\theta}(x,x)-k_{\mathcal{D}}(x)^T(\tilde{K}_\theta(\mathcal{X}))^{-1}k_{\mathcal{D}}(x),
\end{align}
 where $\tilde{K}_\theta(\mathcal{X})={K}_\theta(\mathcal{X})+\sigma^2I_M$ and $k_{\mathcal{D}}(x)$ is the $M \times 1$ covariance vector
$
      k_{\mathcal{D}}(x)=[k_{\theta}(x,x_1),\hdots,k_{\theta}(x,x_M)]^T.
$

We will also require the evidence or \textit{marginal likelihood} of the \textit{output labels},
\begin{equation}
    p_{\theta}({\mathcal{Y}}|\mathcal{X})=\int p_\theta(\mathrm{t}(\mathcal{X})=t)p(\mathcal{Y}|\mathrm{t}(\mathcal{X})=t)dt,
\end{equation}
the log of which can be obtained in closed form for the Gaussian likelihood as \cite{rasmussen2003gaussian}
\begin{align}
    &\ln{p_{\theta}({\mathcal{Y}}=y|\mathcal{X})}\nonumber \\=&-\frac{1}{2}(y-\mu_\theta(\mathcal{X}))^T(\tilde{K}_\theta(\mathcal{X}))^{-1}(y-\mu_\theta(\mathcal{X}))\nonumber \\&-\frac{1}{2}\ln{|\tilde{K}_\theta(\mathcal{X})|}-\frac{M}{2}\ln{2\pi}.\label{eq:evidence}
\end{align}
\subsection{Meta-learning the GP Prior (PACOH-GP)}
\label{ssec:metagp}
In GP, the hyperparameter vector $\theta \in \Theta$ that describes the mean and kernel functions of the GP in \eqref{eq:prior_1} is fixed \emph{a priori}, possibly using cross-validation. In contrast, the meta-learning approach introduced in \cite{rothfuss2020pacoh} -- termed PACOH-GP -- aims to automatically infer the hyperparameters $\theta$ of the GP prior by observing data from tasks with similar statistical properties \cite{rothfuss2020pacoh}. Note that, the kernel function could be parametrized as
\begin{equation}
k_\theta(x,x')=\frac{1}{2}\exp{(-||\Phi_{\theta}(x)-\Phi_{\theta}(x')||_2^2}),
\end{equation}
where $\Phi_{\theta}(\cdot)$ is a parametric function, typically a deep neural network, with $\theta$ constituting its weight and biases.

Following the setting of Baxter \cite{baxter2000model}, the tasks observed by a meta-learner are assumed to be sampled i.i.d from a \textit{task environment}, that defines a distribution $P_{T}$ over the space of tasks. Precisely, for each $i$th observed task $\tau_i$, we sample a data distribution $P_i \sim P_{T}$ from the task environment, and the corresponding data set $\mathcal{D}_i=(\mathcal{X}_i,\mathcal{Y}_i)=\{x_{i,m},y_{i,m}\}_{m=1}^{M_i}$ of $M_i$ samples are generated i.i.d. according to the unknown population distribution $P_i$. The data set generated from observing $N$ meta-training tasks  constitutes the \textit{meta-training set} $\mathcal{D}_{1:N}=(\mathcal{D}_1,\hdots,\mathcal{D}_N)$. 

At test time, the meta-learner is given data from a \textit{meta-test task} with unknown population distribution $P$ drawn from the task environment $P_T$. We denote $\mathcal{D}=(\mathcal{X},\mathcal{Y})$ as the $M$-sample \textit{meta-test training data set} generated i.i.d according to $P$, and $(x,y)$ as the test data sample from the meta-test task. 

Following the non-parametric Bayesian model described in the previous subsection, we model each $i$th observed task through a random scalar function $\trm_i(x)$. Importantly, tasks share the same GP prior $\mathcal{GP}(\mu_{\theta}(x),k_{\theta}(x,x'))$. Therefore, transfer of information among data points of different tasks takes place through the hyperparameter $\theta$.

The goal of meta-learning is  to infer the shared hyperparameter $\theta$ of the GP prior, using the meta-training data $\mathcal{D}_{1:N}$, for use on a new, previously unseen \textit{meta-test} task. The meta-test task is modelled by a random function $\trm(\cdot) \sim \mathcal{GP}(\mu_{\theta}(x),k_{\theta}(x,x'))$ that share the same hyperparameter $\theta$ as the meta-training tasks. 


 
The shared hyperparameter $\theta$ is endowed with a \textit{hyper-prior} distribution $p(\theta)$, and the meta-learner uses the \textit{meta-training} data set $\mathcal{D}_{1:N}$ to update the hyper-prior $p(\theta)$ to a \textit{hyper-posterior} $q(\theta|\mathcal{D}_{1:N})$.
%
This is done by minimizing a free energy metric. Specifically, the \textit{meta-training loss} of hyperparameter vector $\theta \in \Theta$ incurred on the meta-training set $\mathcal{D}_{1:N}$ is defined as
 \begin{align}
     \mathcal{L}(\theta,\mathcal{D}_{1:N})=\frac{1}{N} \sum_{i=1}^N \frac{-\log p_{\theta}(\mathcal{Y}_i|\mathcal{X}_i)}{M_i},
     \vspace{-0.4cm}
 \end{align}which is the empirical average of the negative log-marginal likelihood $p_{\theta}(\mathcal{Y}_i|\mathcal{X}_i)$ across the meta-training tasks. This is defined as in \eqref{eq:evidence} for a Gaussian likelihood. The hyper-posterior $q(\theta|\mathcal{D}_{1:N})$ is then optimized so as to minimize the \textit{free energy objective} \cite{jose2021free}, \begin{align} \mathcal{F}(q)=\mathbb{E}_{ q(\theta|\mathcal{D}_{1:N})}[\mathcal{L}(\theta,\mathcal{D}_{1:N})]  +\gamma^{-1} D(q(\theta|\mathcal{D}_{1:N})||p(\theta)), \label{eq:freeenergy_metalearning} \end{align} where  $\gamma>0$ is a \textit{temperature} parameter, and $D(p||q)$ denotes the KL divergence between the distributions $p$ and $q$.
 
 The meta-free energy function $\mathcal{F}(q)$ is the sum of: $(a)$ the \textit{average meta-training loss} for a randomly drawn hyperparameter $\theta \sim q(\theta|\mathcal{D}_{1:N})$; and $(b)$ the KL-divergence term between the hyper-posterior $q(\theta|\mathcal{D}_{1:N})$ and the hyper-prior $p(\theta)$, which serves as a regularization on the meta-level.
 Let $\gamma^{-1}=(1/N+1/\widetilde{M})$, with $\widetilde{M}=(N^{-1}\sum_{n=1}^{N}M_n^{-1})^{-1}$ serve as the harmonic mean, the function $\mathcal{F}(q)$ in \eqref{eq:freeenergy_metalearning} corresponds to an upper bound (neglecting constant terms) on the population test log-loss obtained via PAC-Bayes analysis \cite{rothfuss2020pacoh}. The optimization problem in \eqref{eq:freeenergy_metalearning} also corresponds to a form of information risk minimization (IRM) \cite{zhang2006information} and generalized Bayesian meta-learning \cite{rothfuss2020pacoh}. 
 
 The minimizing solution
can be obtained in closed form as the \emph{Gibbs hyper-posterior}  \cite{rothfuss2020pacoh}
\begin{equation}
 q^{PACOH-GP}(\theta|\mathcal{D}_{1:N})\propto p(\theta)\exp\Bigl(-\gamma \mathcal{L}(\theta,\mathcal{D}_{1:N}) \Bigr).
\end{equation}During meta-testing, we evaluate the average predictive distribution as
\begin{align}
    \mathbb{E}_{q^{PACOH-GP}(\theta|\mathcal{D}_{1:N})}[p_{\theta}(\mathrm{t(x)}=t|\mathcal{D})], \label{eq:average_prediction}
\end{align} of the meta-test task function $\trm(\cdot)$ at its test input $x$.


\section{Transfer Meta-learning the GP Prior}
\label{sec:transfer}
In this section, we introduce and formulate the problem of  \textit{transfer} meta-learning the GP prior, inspired by the recent theoretical work in \cite{jose2020transfer}.  As explained in the previous section, in conventional meta-learning, the meta-training and meta-test tasks belong to the same \textit{task environment} in the sense that the population distributions for each task are drawn from the same task distribution $P_T$. In contrast, transfer meta-learning concerns  settings in which the observed meta-training tasks belong to a \textit{source task environment}, while the meta-test task belongs to a different \textit{target task environment}. The task distributions for the two environments are respectively denoted as $P_{T}^{S}$ and $P_{T}^{T}$.

In this work, we assume that the transfer meta-learner, in addition to data from the source task environment, observes data from  a limited number of tasks from the target task environment. As such, the meta-training set $\mathcal{D}_{1:N}=(\mathcal{D}_1,\hdots,\mathcal{D}_N)$ comprises of $N$ data sets, of which a subset of $\beta N$ data sets, for $\beta \in [0,1]$, correspond to tasks from the source task environment, and the remaining from the target task environment. Accordingly, for each task $\tau_i$, $i=1,\hdots,\beta N$, a data distribution $P_i$ is sampled from the source task environment $P_{T}^{S}$, with the corresponding $M_i$-sample training data  generated i.i.d. according to $P_i$. For $i=\beta N+1,\hdots, N$, each task $\tau_i$ samples a data distribution $P_i$ from the target task environment $P_{T}^{T}$.
The meta-test task is drawn from the target task environment. Let $\mathcal{D}=(\mathcal{X},\mathcal{Y})$ denote the $M$-sample meta-test training data, and $(x,y)$ a meta-test test data point.

The goal of transfer meta-learning is to use the observed meta-training data set $\mathcal{D}_{1:N}$ to infer a hyperparameter vector $\theta$ of the GP prior for use on a new meta-test task. 
As in \cite{rothfuss2020pacoh}, the transfer meta-learner uses the meta-training set $\mathcal{D}_{1:N}$ to update the hyper-prior $p(\theta)$ to a hyperposterior $q(\theta|\mathcal{D}_{1:N})$.

To this end, the transfer meta-learner considers the following \textit{weighted average meta-training loss},
\begin{align}
   \hspace{-0.1cm} \bar{\mathcal{L}}(\theta,\mathcal{D}_{1:N})\hspace{-0.1cm}=\hspace{-0.1cm}\alpha \mathcal{L}_s(\theta,\mathcal{D}_{1:\beta N})\hspace{-0.1cm}+\hspace{-0.1cm}(1-\alpha)\mathcal{L}_t(\theta,\mathcal{D}_{\beta N+1:N}), \label{eq:metatrainingloss}
\end{align}for $\alpha \in [0,1]$, which is a convex combination of the training loss
\begin{equation*}
    {\mathcal{L}}_s(\theta,\mathcal{D}_{1:\beta N}) = \frac{1}{\beta N} \sum_{i=1}^{\beta N} \frac{-\log p_{\theta}(\mathcal{Y}_i|\mathcal{X}_i)}{M_i}
\end{equation*}
evaluated on data from the source environment and of the training loss computed on data from the target environment
\begin{equation*}
    \mathcal{L}_t(\theta,\mathcal{D}_{\beta N+1:N})  = \frac{1}{(1-\beta) N} \sum_{i=\beta N+1}^{ N} \frac{-\log p_{\theta}(\mathcal{Y}_i|\mathcal{X}_i)}{M_i} .
\end{equation*}
In the proposed \textbf{weighted free energy minimization with Gaussian Processes (WFEM-GP)}, the target hyperposterior $q(\theta|\mathcal{D}_{1:N})$ is optimized so as to minimize the \emph{weighted free energy functional} 
\begin{equation}
    \mathcal{F}(q)=\mathbb{E}_{ q(\theta|\mathcal{D}_{1:N})}[\bar{\mathcal{L}}(\theta,\mathcal{D}_{1:N})]  +\gamma^{-1} D(q(\theta|\mathcal{D}_{1:N})||p(\theta)). \label{eq:weightedmetafree}
\end{equation}

Setting $M_i=M$, for $i=1,\hdots, N$ and $\gamma^{-1}=(1/N+1/M)$, the free energy functional in \eqref{eq:weightedmetafree} (neglecting constant terms) provides a PAC-Bayesian upper bound on the population test log-loss \cite{jose2020transfer}. 
Similar to (13), the minimizing solution is obtained as the \emph{Gibbs hyper-posterior}
\begin{equation}
    q^{WFEM-GP}(\theta|\mathcal{D}_{1:N})\propto p(\theta)\exp{\left[-\gamma \bar{\mathcal{L}}(\theta,\mathcal{D}_{1:N})\right]}. \label{eq:transfer_gibbshyperposterior}
\end{equation} Finally, the hyper-posterior $q^{WFEM-GP}(\theta|\mathcal{D}_{1:N})$ is used in lieu of the corresponding PACOH-GP hyperposterior in \eqref{eq:average_prediction} in order to define the predictive distribution.

In practice, for both PACOH-GP and WFEM-GP, the expectations in the predictive distributions \eqref{eq:average_prediction} need to be approximated. As detailed in the supplementary materials \ref{app:approximationschemes}, this can be done by evaluating the maximum of the hyperposteriors, and by plugging this value into the predictive distribution $p_{\theta}(\mathrm{t(x)}=t|\mathcal{D})$ -- an approach we refer to as \emph{maximum a posteriori (MAP)}. Alternatively one can use an average obtained via the particle-based inference through \emph{Stein Variational Gradient Descent (SVGD)} \cite{liu2016stein}. 

\section{Experiment Setting}
\label{sec:exp}
In this section, we detail our experimental setting and compare WFEM-GP and PACOH-GP on synthetic and real-world datasets.

\subsection{Sinusoidal Regression}
\label{subsec:sinusoid}
We first demonstrate the advantage of WFEM-GP over GP and PACOH-GP by considering a sinusoidal regression problem. Towards this, we first describe the data generation process for source and target environment. For each task, the input $x$ is drawn from a uniform distribution $\mathrm{Unif(-5,5)}$. The output $y$ corresponding to an input $x$ is obtained as $y \sim \mathcal{N}(y|f(x),0.1^2)$, where
\begin{equation}
    f(x)= \mathrm{a} x+ \mathrm{b} \sin{(1.5(x-\mathrm{c}))}+\mathrm{d}, \label{eq:groundtruth}
\end{equation}
and the scalars $\mathrm{a}, \mathrm{b}, \mathrm{c}$ and $\mathrm{d}$ characterize a given task. The source (or target) task environment defines a joint distribution over the parameters $(\mathrm{a}, \mathrm{b}, \mathrm{c},\mathrm{d})$. Specifically, each task from the source task environment is sampled as
\begin{align}
    &\mathrm{a}\sim\mathcal{N}(0.5,0.2^2),    \mathrm{b}\sim\mathcal{U}(0.7,1.3),\nonumber \\&\mathrm{c}\sim\mathcal{N}(\mu_c,0.1^2),
    \mathrm{d}\sim\mathcal{N}(5.0,0.1^2), \label{eq:taskdistribution_source}
\end{align}
where $\mu_c \in \mathbb{R}$ denotes the mean value of the parameter $\mathrm{c}$.
We consider the target task environment to follow the same distributions for parameters $\mathrm{a},\mathrm{b}$ and $\mathrm{d}$ as in \eqref{eq:taskdistribution_source}, while the parameter $\mathrm{c}$ is distributed as
$
    \mathrm{c} \sim \mathcal{N}(\mu'_c,0.1^2),
$ with a mean $\mu'_c$ distinct from $\mu_c$.
We assume a Gaussian likelihood for each $i$th task as $p(y|t)=\mathcal{N}(y|t,\sigma^2)$.
\vspace{-0.2cm}
\begin{figure}[htb]

  \centering
  \centerline{\includegraphics[scale=0.24,trim=0in 0in 0in 0in,clip]{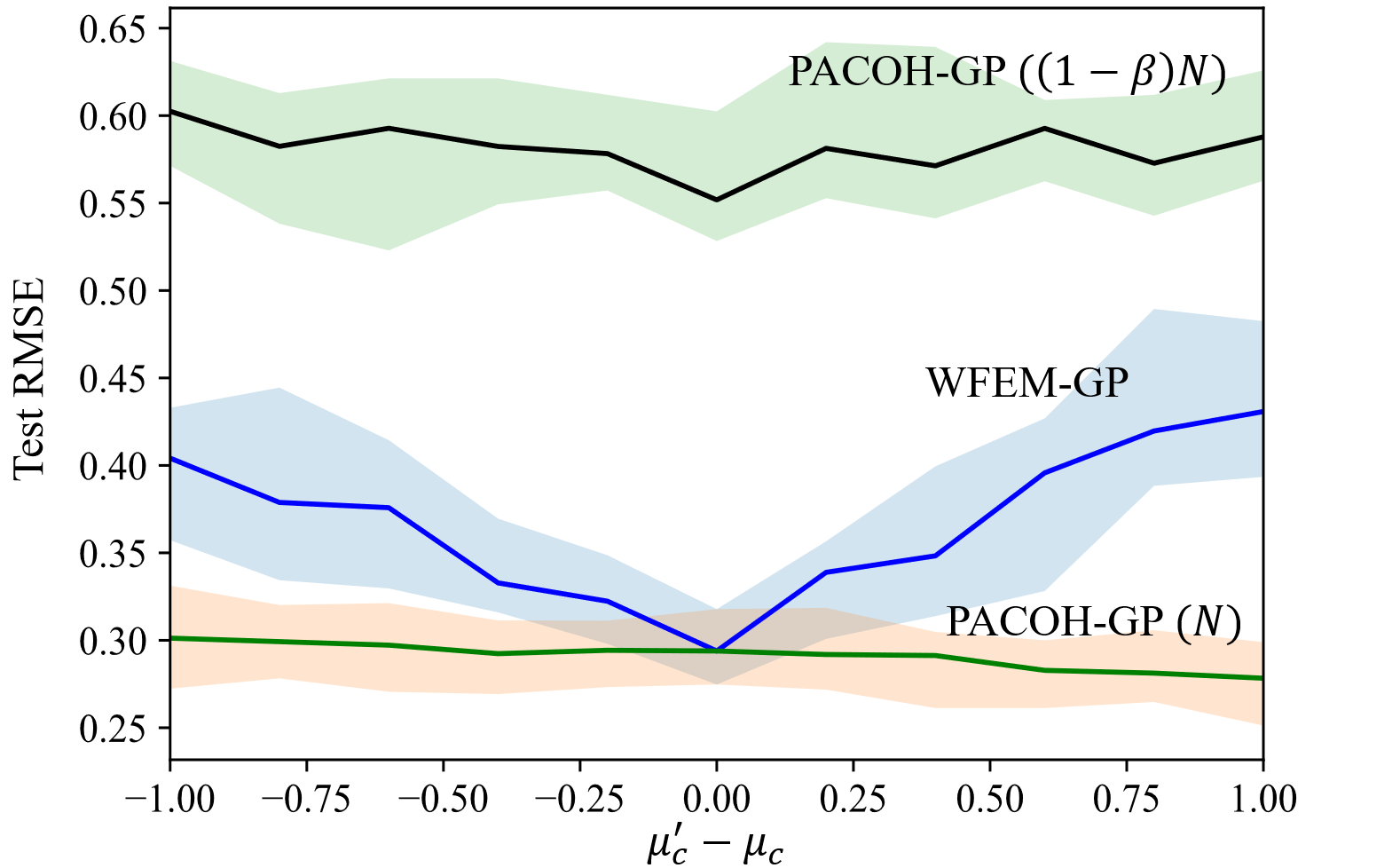}}
  \vspace{-0.3cm}
%
%
\caption{Average test root mean square error (RMSE) under three schemes -- PACOH-GP with $N$ tasks and  with $(1-\beta)N$ tasks from the target environment and WFEM-GP --  as a function of the deviation $\mu'_c-\mu_c$ of the target task environment from the source task environment with fixed $\mu_c=0$.
}  
\label{fig:res1}
%
\end{figure}

In Figure 1, we compare the performance  of WFEM-GP with that of PACOH-GP that uses $N$ tasks or $(1-\beta)N$ tasks from target environment. We vary the deviation $\mu_c'-\mu_c$ of the mean $\mu_c'$ of parameter $\mathrm{c}$ in the target environment from a fixed mean $\mu_c$ of the source environment. We set $\sigma=0.1$, $\alpha=\beta=0.5$, number of tasks $N=30$, and number of samples per task $M_i=5$. We adopt the MAP approximation for all the learning schemes.

WFEM-GP is seen to outperform the PACOH-GP scheme that uses only $(1-\beta)N$ tasks from the target environment. This suggests that data from the source task environment can be utilized during meta-training to improve the performance on test tasks from the target environment. We also benchmark the performance of WFEM-GP against the ideal performance of a meta-learner trained on $N$ target tasks. It can be seen that the transfer meta-learner, which is trained on limited number of tasks from the target environment performs close to this ideal reference, and that it coincides with it when the deviation in task distributions is zero, i.e, when source and target task environments are the same.



\begin{figure}[htb]

  \centering
  \centerline{\includegraphics[scale=0.29]{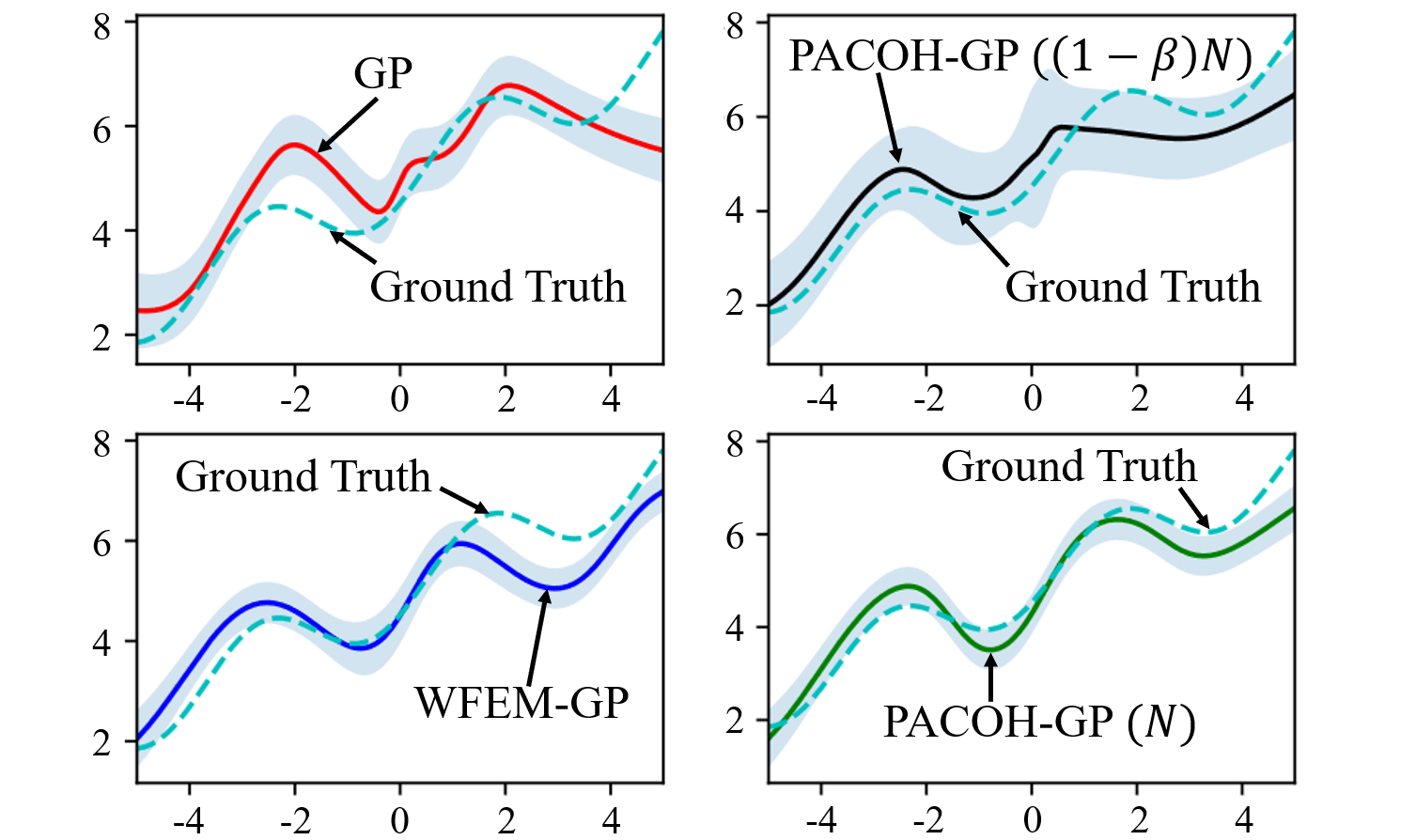}}
  \vspace{-0.3cm}
  \caption{Comparison of posterior predictions under the four schemes --  GP, PACOH-GP with $N$ tasks and  with $(1-\beta)N$ tasks from the target environment and WFEM-GP -- against the ground-truth regression function.
  }
\label{fig:res2}
\end{figure}

In Figure~\ref{fig:res2}, we compare the posterior predictions of the three schemes introduced above, along with GP, against the ground truth regression function. We set $\sigma=0.1$, $\alpha=\beta=0.2$, $\mu_c'-\mu_c=0.5$, number of tasks $N=30$ and number of samples per task $M_i=5$. We adopt the MAP for all learning schemes. The dashed line represents the ground truth regression line in \eqref{eq:groundtruth}. The performance of WFEM-GP is again comparable to the best achievable performance of a meta-learner trained on $N$ tasks from the target environment.

In Figure 3, we compare the performance of the four schemes outlined above as a function of the fraction $\beta$ of tasks from the source task environment. We set $\sigma=0.1$, $\alpha=\beta$, $\mu_c'-\mu_c=0.75$, $N=30$ and  $M_i=5$. When $\beta=0$, only tasks from the target environment are available for meta-training, and hence the PACOH-GP and WFEM-GP schemes coincide. At the other extreme, when $\beta=1$, i.e., only tasks from the source environment are available for meta-training, PACOH-GP using $(1-\beta)N$ target tasks coincides with GP as the two schemes share the same initial hyperparameter $\theta$. In general, as $\beta$ increases, WFEM-GP increasingly deviates from the ideal performance of the meta-learner trained on $N$ target tasks, while clearly outperforming PACOH-GP with $(1-\beta)N$ target tasks.
\begin{figure}[htb]

  \centering
  \centerline{\includegraphics[scale=0.23]{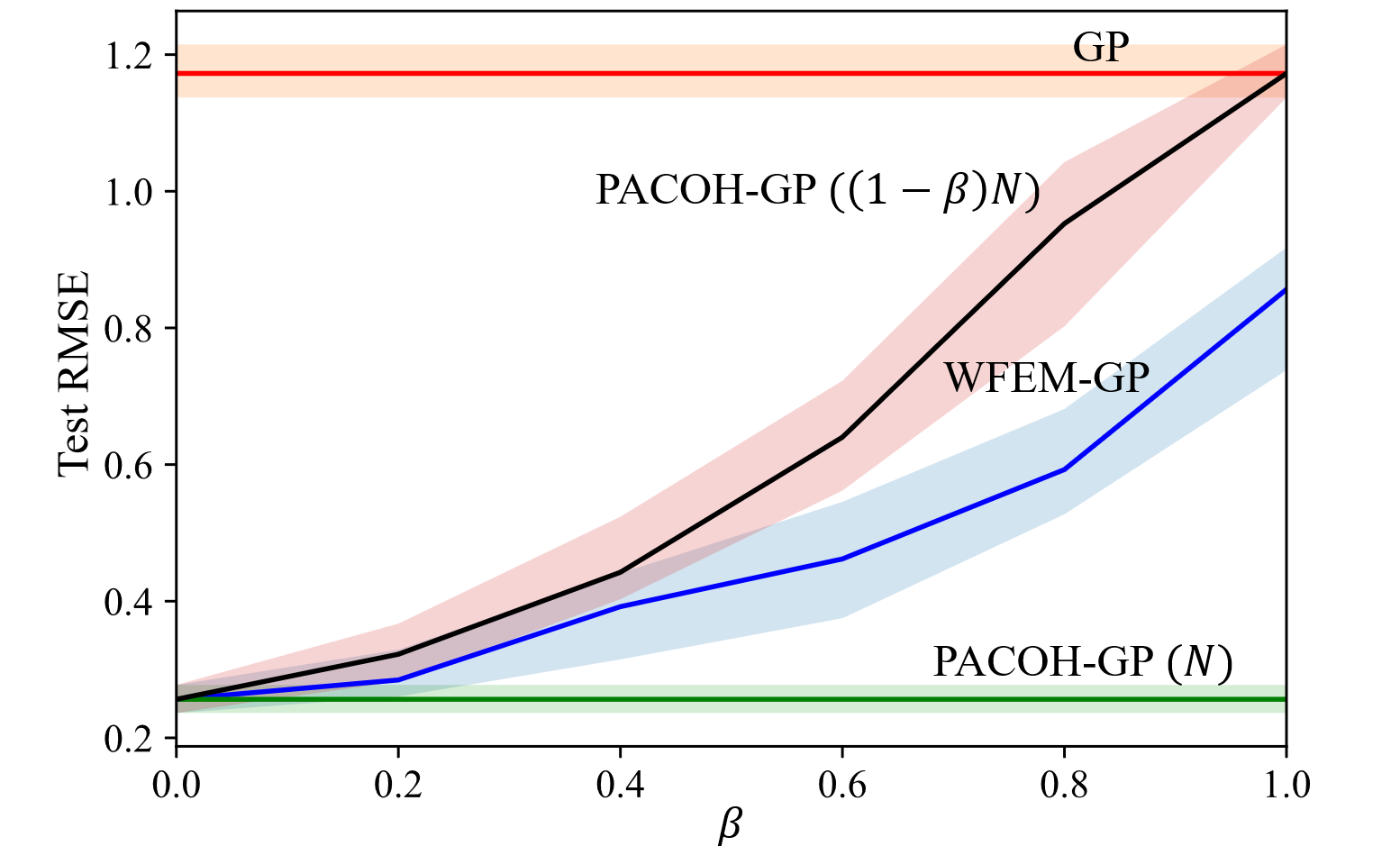}}
  \caption{Average test RMSE under four schemes -- GP, PACOH-GP with $N$ tasks and  with $(1-\beta)N$ tasks from the target environment and WFEM-GP --  as a function of  $\beta$ with fixed $\mu'_c-\mu_c=0.75$. 
  } 
\label{fig:res3}
\end{figure}

\subsection{GP Classification}\label{sec:GP_classification}
In this section, we consider binary classification experiments on real-world data sets using GP. Unlike the GP regression example considered Section~\ref{subsec:sinusoid}, employing GP for classification is not straightforward since the likelihood $p(y|t)$ is non-Gaussian. To this end, we assume a logistic regression model with $p(y=+1|t)=\sigma(t)$, where $\sigma(a)=(1+\exp(-a))^{-1}$ is the sigmoid function. Classification is then performed using a Laplace approximation-based classifier \cite{rasmussen2003gaussian}. We refer the readers to Appendix~\ref{app:GP_classification} for more details.

We evaluate the performance of the proposed transfer meta-learner in \eqref{eq:transfer_gibbshyperposterior} using  standard few-shot classification datasets, namely mini-ImageNet serving as source task environment and CUB for the target task environment. The mini-ImageNet is composed of 100 classes selected from ImageNet randomly, and each class has 600 images, which are resized to 84×84 pixels for fast training and inference \cite{NIPS2016_90e13578}. CUB is composed of 11788 images over 200 birds classes, which are also resized to 84×84 pixels \cite{WahCUB_200_2011}. 

We conduct 2-way 5-shot binary classification experiments based on above datasets. Precisely, the data set for each task from the source task environment (mini-ImageNet) is obtained by first selecting 2 classes at random, and then randomly sampling  5 images for each class from mini-ImageNet dataset. The training data set from target task environment is similarly chosen from the CUB data set.
For testing tasks, we sample randomly 15 images from each class. 

In Figure 4, we compare the performance of WFEM-GP with the three benchmark schemes - GP, PACOH-GP with $(1-\beta)N$ tasks and with $N$ tasks from target task environment -- as a function of $\alpha$. We plot the performances under both MAP and SVGD schemes. 
Other parameters are set as $N=20$, $M_i=5$ and $\beta=0.5$. Confirming the results in \cite{rothfuss2020pacoh}, SVGD outperforms MAP for all learning schemes. Moreover, WFEM-GP is observed to outperform GP, PACOH-GP with $(1-\beta)N$ target tasks and partially bridge the gap to the ideal PACOH-GP with $N$ target tasks.
\begin{figure}[htb]

  \centering
  \centerline{\includegraphics[scale=0.26]{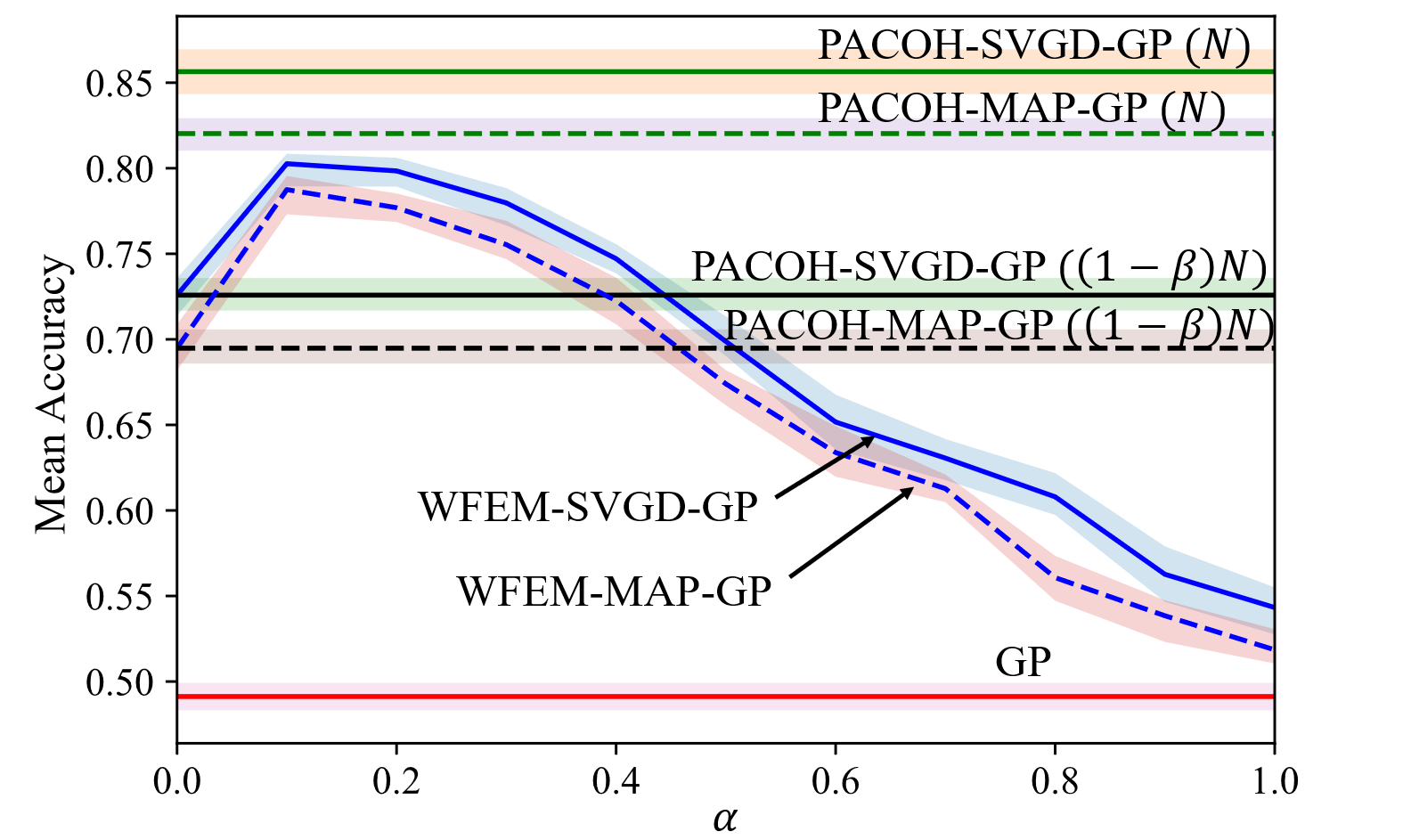}}
  \vspace{-0.5cm}
  \caption{Comparison of mean accuracy between MAP and SVGD of GP, PACOH-GP with $N$ tasks and with $(1-\beta)N$ tasks from the target environment and WFME-GP against varying weight parameter $\alpha$ of the meta-training loss.}
\label{fig:res5}
\vspace{-0.4cm}
\end{figure}
\bibliographystyle{IEEEbib}
\bibliography{ref}

\newpage
\appendix
\section{Approximation Schemes for Gibbs Hyperposterior $ q^{WFEM-FP}(\theta|\mathcal{D}_{1:N})$ }\label{app:approximationschemes}
In this section, we describe two tractable schemes for approximating the Gibbs hyperposterior \eqref{eq:transfer_gibbshyperposterior}.

\noindent \emph{Maximum A Posteriori (MAP) Estimate:} The MAP estimate approximates the Gibbs hyperposterior 
$q^{WFEM-GP}(\theta|\mathcal{D}_{1:N})$ by a Dirac measure centered at its mode
\begin{align}
    \theta^{*}=\arg \max_{\theta \in \Theta} q^{WFEM-GP}(\theta|\mathcal{D}_{1:N}).
\end{align}
The mode $\theta^{*}$ can be evaluated equivalently as the solution to the following optimization problem,
\begin{equation}
    \theta^*=\mathop{\arg\min_{\theta \in \Theta}}\left\{-\log{p(\theta)}+\gamma^{-1}\bar{\mathcal{L}}(\theta,\mathcal{D}_{1:N})\right\}. \label{eq:thetastar}
\end{equation}
With the choice of a zero mean isotropic Gaussian distribution as the hyper-prior, i.e., $p(\theta)\sim \mathcal{N}(0,\sigma_{\mathcal{P}}^2I)$, \eqref{eq:thetastar} results in
\begin{equation}
\begin{split}
   \arg \min_{\theta \in \Theta} J^{MAP}(\theta)=\gamma^{-1}\bar{\mathcal{L}}(\theta,\mathcal{D}_{1:N})+\frac{1}{2\sigma_{\mathcal{P}}^2}||\theta||^2,
\end{split}
\end{equation}
where the objective $J^{MAP}(\theta)$ consists of a sum of meta-training loss \eqref{eq:metatrainingloss} and an $L2$-regularization term.

To optimize $J^{MAP}(\theta)$, we use mini-batch gradient descent in training, where the mini-batches are sampled at meta-level (i.e., we sample mini-batches of tasks and use all data points of the corresponding tasks to compute the gradients of $J^{MAP}(\theta)$.)

\noindent \emph{Stein Variational Gradient Descent (SVGD):}
SVGD is a general purpose variational inference algorithm that aims to minimize the KL divergence, $D(\hat{q}(\theta)||q^{WFEM-GP}(\theta|\mathcal{D}_{1:N}))$, to the target distribution $q^{WFEM-GP}(\theta|\mathcal{D}_{1:N})$, over non-parametric distributions $\hat{q}(\theta)$. The distribution $\hat{q}(\theta)$ is represented by a collection of particles $\{\theta_1, \theta_2,...,\theta_K\}$, which can in turn be used to approximate $\hat{q}(\theta)$ via a Kernel Density Estimator (KDE) \cite{bishop2006pattern}. 

For the SVGD approximation of $q^{WFEM-GP}(\theta|\mathcal{D}_{1:N})$, we start by sampling $K$ particles $\{\theta_1, \theta_2,...,\theta_K\}$ from the hyper-prior $p(\theta)$. The particles are then iteratively transported to minimize the KL divergence $D(\hat{q}(\theta)||q^{WFEM-GP}(\theta|\mathcal{D}_{1:N}))$ via a form of functional gradient descent on a reproducing kernel Hilbert space (RKHS), induced by a kernel function $\tilde{k}(\cdot,\cdot)$. Specifically, we choose a squared exponential kernel, $$\tilde{k}(\theta,\theta')=\exp \biggl( -\frac{||\theta-\theta'||^2}{2l}\biggr),$$
with $l$ denoting the fixed, length hyperparameter.
Consequently, the SVGD update at iteration $l$ is given as
\begin{align}
    \theta_k^{[l]} \leftarrow \theta_k^{[l-1]} +\epsilon \Phi(\theta_k^{[l-1]}),
\end{align} where
\begin{align}
    &\Phi(\theta_k^{[l-1]})= \nonumber \\& \frac{1}{K}\sum_{j=1}^K \Bigl[ \tilde{k}(\theta_j^{[l-1]},\theta_k^{[l-1]})\nabla_{\theta_j^{[l-1]}} \log q^{WFEM-GP}(\theta_j^{[l-1]}|\mathcal{D}_{1:N})\nonumber \\&+\nabla_{\theta_j^{[l-1]}} \tilde{k}(\theta_j^{[l-1]},\theta_k^{[l-1]})\Bigr]
\end{align} for each particle $k=1,\hdots,K$.
Moreover, it has been shown in \cite{liu2016stein} that in the asymptotic limit as $K \rightarrow \infty$, the empirical distribution encoded by the particles $\{ \theta_1, \hdots, \theta_K\}$ converges to the true target distribution $q^{WFEM-GP}(\theta|\mathcal{D}_{1:N}).$

To estimate the score function $\nabla_{\theta_j^{[l-1]}} $ $\log q^{WFEM-GP}(\theta_j^{[l-1]}|\mathcal{D}_{1:N})$, we use a mini-batch of $n$ meta-training data sets with $\mathcal{D}_1,\hdots,\mathcal{D}_{\beta n}$ data from source environment data set and $\mathcal{D}_{\beta n+1},\hdots,\mathcal{D}_{n}$ from target environment data set. Using this, the score function is approximated as
\begin{align}
& \nabla_{\theta_j^{[l-1]}} \log q^{WFEM-GP}(\theta_j^{[l-1]}|\mathcal{D}_{1:N}) \nonumber\\
& \approx \frac{N}{n} \gamma \biggl[\alpha \sum_{i=1}^{\beta n} \frac{1}{m_i} \nabla_{\theta_j^{[l-1]}} \log p_{\theta_j^{[l-1]}}(\mathcal{Y}_i|\mathcal{X}_i) \nonumber \\&+ (1-\alpha)\sum_{i=\beta n+1}^{n} \frac{1}{m_i} \nabla_{\theta_j^{[l-1]}} \log p_{\theta_j^{[l-1]}}(\mathcal{Y}_i|\mathcal{X}_i)\biggr] \nonumber \\&+ \nabla_{\theta_j^{[l-1]}} \log p(\theta_j^{[l-1]}).
\end{align}


\section{Meta-Testing of WFEM-GP under MAP and SVGD}\label{app:metatesting}
During meta-testing, the hyperparameter meta-learned using WFEM-GP is used to initialize the GP prior of a meta-test task. The GP is subsequently fitted to the meta-test training data set $\mathcal{D}=(\mathcal{X},\mathcal{Y})$ of $M$ samples to yield a GP posterior. We assume that a test data set, $\mathcal{D}^*=\{(x_m^*,y_m^*)_{m=1}^M\}$, is also available, independent of the training data set $\mathcal{D}$. The performance of the GP posterior for the meta-test task is then evaluated on $\mathcal{D}^{*}$. 

To this end, corresponding to each test input $x_m^{*}$, for $m=1,\hdots, M$,  we evaluate the average predictive posterior distribution $\mathbb{E}_{q^{WFEM-GP}_{\theta|\mathcal{D}_{1:N}}}[p_{\theta}(\trm(x_m^{*})|\mathcal{D})]$ defined in \eqref{eq:average_prediction}.
Under the MAP scheme, which yields a point estimate $\theta^{*}$ of the Gibbs hyperposterior $q^{WFEM-GP}(\theta|\mathcal{D}_{1:N})$,  the average predictive posterior distribution corresponds to 
$p_{\theta^{*}}(\trm(x_m^{*})|\mathcal{D})$. In contrast, SVGD scheme outputs $K$ particles, $\theta_k\sim q^{WFEM-GP}(\theta|\mathcal{D}_{1:N})$, for $k=1,\hdots, K$. These can be used to approximate the average predictive posterior distribution as \begin{equation}
\begin{split}
    &\mathbb{E}_{\theta\sim q^{WFEM-GP}(\theta|\mathcal{D}_{1:N})}[p_{\theta}(\mathrm{t}(x_m^*)|{\mathcal{D}})] \\&\approx\frac{1}{K}\sum_{k=0}^K p_{\theta_k}(\mathrm{t}(x_m^*)|{\mathcal{D}}). \label{eq:average}
\end{split}
\end{equation} It is easy to see that for $K=1$ and $\theta_k=\theta^{*}$, the SVGD scheme coincides with MAP.

To evaluate the predictive performance for regression experiment, we use average root mean square error (RMSE) as the metric, which can be computed as follows. For each $\theta_k$ in the SVGD scheme (or $\theta^{*}$ for MAP scheme), we consider the mean prediction of $p_{\theta_k}(\mathrm{t}(x_m^*)|{\mathcal{D}})$ as $\hat{y}(\theta_k,x_m^{*})=\mathbb{E}_{p_{\theta_k}(\mathrm{t}(x_m^*)|{\mathcal{D}})}[t]$. Subsequently, the mean prediction corresponding to \eqref{eq:average} evaluates as
\begin{align}
    \hat{y}(x_m^{*})=\frac{1}{K}\sum_{k=0}^K \hat{y}(\theta_k,x_m^{*}).
\end{align}
The average root mean squared error (RMSE) is then computed as
\begin{equation}
   \mathrm{RMSE} =\sqrt{\frac{1}{M}\sum_{m=1}^M( \hat{y}(x_m^{*})-y_m^*)^2}.
\end{equation}

In our experiments, we set the number of SVGD particles to be $K=10$.

To evaluate the predictive performance for classification, we adopt mean accuray as the metric, which can be computed as follows. In each dataset, we sum the absolute difference between prediction $\hat{y}(x_m^*)$ and label $y_m^*$, then compute the mean accuracy as
\begin{align}
    \mathrm{Mean Accuracy}=1-\frac{1}{M}\sum_{m=1}^{M}|\hat{y}(x_m^*)-y_m^*|
\end{align}


\section{Laplace Approximation-Based GP Binary Classifier }\label{app:GP_classification}
In this section, we review the Laplace approximation based implementation of the GP binary classifier \cite{rasmussen2003gaussian} which trains on an input data set $\mathcal{D}=(\mathcal{X},\mathcal{Y})$. We outline the key steps here and refer the readers to \cite{rasmussen2003gaussian} for more details.

As detailed in Section~\ref{sec:GP_classification}, for the binary classification problem using GP, we assume a logistic regression model with $p(y=+1|t)=\sigma(t)$, where $\sigma(a)=(1+\exp(-a))^{-1}$ is the sigmoid function. 
Note that the function $t$ acts as a latent function in describing the likelihood $p(y=+1|t)$:  inferring $t$ does not yield the required predictions as in the regression problem, but has to be combined with a deterministic sigmoid function. 

In GP classification, the latent function $\trm(\cdot)$ is assumed to be random and endowed with a GP prior. As such, corresponding to an observed input data set $\mathcal{X}$, the distribution $p_{\theta}(\trm(\mathcal{X}))$ defines a GP prior over the output of the latent function $\trm(\mathcal{X})$, and  $$p(\mathcal{Y}=y|\trm(\mathcal{X})=t)=\prod_{m=1}^M\sigma(t_m)^{y_m}(1-\sigma(t_m))^{(1-y_m)},$$ with $y=[y_1,\hdots,y_M]^T$ and $t=[t_1,\hdots,t_M]^T$ describes the
likelihood of the data set. In contrast to GP regression, the above likelihood is non-Gaussian. The GP posterior resulting from fitting the data set $\mathcal{D}=(\mathcal{X},\mathcal{Y})$ on the GP prior is then obtained as \begin{equation}
p_{\theta}(\trm(\mathcal{X})=t|\mathcal{D})=\frac{p(\mathcal{Y}|\trm(\mathcal{X})=t)p_{\theta}(\trm(\mathcal{X})=t)}{p_{\theta}(\mathcal{Y}|\mathcal{X})}. \label{eq:posterior}
\end{equation}

Inference is done in two steps: in the first step, we evaluate the distribution of the output of the latent function $\trm(\cdot)$ with respect to a test input $x$ as 
\begin{equation}
p_{\theta}(\trm(x)|\mathcal{D})= \int p_{\theta}(\trm(x)|\trm(\mathcal{X})=t) p_{\theta}(\trm(\mathcal{X})=t|\mathcal{D})dt, \label{eq:int1}
\end{equation} where the conditional distribution $p_{\theta}(\trm(x)|\trm(\mathcal{X})=t)$ is Gaussian and can be evaluated directly as in \cite[Equation 2.19]{rasmussen2003gaussian}.
 The distribution of the test output of the latent function is subsequently used to make prediction as
\begin{equation}
    p(y=1|\mathcal{D}, x)=\int \sigma(t)p_{\theta}(\trm(x)=t|\mathcal{D})dt. \label{eq:int2}
\end{equation}
The integrals in \eqref{eq:int1} and \eqref{eq:int2} are intractable due to the non-Gaussianity of the data likelihood. To tackle this, we use a Laplace approximation based classifier obtained via the following steps.

\noindent \emph{Approximation of the posterior $p_{\theta}(\trm(\mathcal{X})|\mathcal{D})$}: The first step is to replace the posterior distribution $p_{\theta}(\trm(\mathcal{X})|\mathcal{D})$  in \eqref{eq:int1}, which is defined as in \eqref{eq:posterior}, using a Laplace approximation \cite{bishop2006pattern}, \begin{align}
  q_{\theta}(\trm(\mathcal{X})=t|\mathcal{D}) \sim \mathcal{N}(t|\hat{t}, \Sigma^{-1}), \label{eq:Laplace}
\end{align} where $\hat{t}=\arg \max_t p_{\theta}(\trm(\mathcal{X})=t|\mathcal{D})$ is the mode of the posterior distribution,  and $\Sigma=-\nabla^2 \log p_{\theta}(\trm(\mathcal{X})|\mathcal{D})|_{\trm(\mathcal{X})=\hat{t}}$ is the Hessian of the negative of the log posterior evaluated at $\hat{t}$.

Computing the mode $\hat{t}$ of the posterior distribution amounts to solving the following equation, 
\begin{align}
    \log{p(\mathcal{Y}|\trm(\mathcal{X})=\hat{t})}-K_{\theta}(\mathcal{X})^{-1}\hat{t}=0,
\end{align} which cannot be directly solved. As such, $\hat{t}$ is obtained using Newton method which iteratively updates an estimate $\tilde{t}$ of the mode as
\begin{equation}
   \tilde{t} \leftarrow (K_{\theta}(\mathcal{X})^{-1}+{W})^{-1}({W}\tilde{t}+\nabla\log p(\mathcal{Y}|\trm(\mathcal{X}))|_{\trm(\mathcal{X})=\tilde{t}}). 
\end{equation} Here, ${W} = -\nabla^{2} \log p(\mathcal{Y}|\trm(\mathcal{X}))|_{\trm(\mathcal{X})=\tilde{t}}$ is the Hessian of the negative log-likelihood evaluated at $\tilde{t}$.

The covariance matrix of Laplace approximation $q_{\theta}(\trm(\mathcal{X})=t|\mathcal{D})$ in \eqref{eq:Laplace}  corresponds to $\Sigma=(K_{\theta}(\mathcal{X})^{-1}+\mathrm{W})$.

\noindent \emph{Approximation of Distribution $p_{\theta}(\trm(x)|\mathcal{D})$:} In the next step, we approximate the test output distribution $p_{\theta}(\trm(x)|\mathcal{D})$  by $q_{\theta}(\trm(x)|\mathcal{D})$, which is obtained by replacing the posterior $p_{\theta}(\trm(\mathcal{X})|\mathcal{D})$ in \eqref{eq:int1} with its Laplace approximation $q_{\theta}(\trm(\mathcal{X})|\mathcal{D})$ obtained in \eqref{eq:Laplace}. This yields that
\begin{equation}
    q_{\theta}(\trm(x)=t|\mathcal{D})=\mathcal{N}(t|\mu(x), s^2(x)),
\end{equation}
where its mean and covariance functions are denoted as
\begin{align}
&\mu(x)=\mu_{\theta}(x)+k_{\mathcal{D}}(x)^{T}\nabla \log p(\mathcal{Y}|\trm(\mathcal{X})=\hat{t}),\\
&s^2(x)= k_{\theta}(x,x)-k_{\mathcal{D}}(x)^T(K_{\theta}(\mathcal{X})+{W}^{-1})^{-1}k_{\mathcal{D}}(x).
\end{align}


\noindent \emph{Computation of Predictive Distribution $p(y=1|\mathcal{D},x)$:} 
Using all the approximations explained above, the predictive distribution $p(y=1|\mathcal{D},x)$ is finally computed as
\begin{equation}
    p(y=1|\mathcal{D},x)\approx \int \sigma(t)q_{\theta}(\trm(x)=t|\mathcal{D})dt.
\end{equation}
The above integral is evaluated by sampling multiple latent function outputs $t_1,\hdots,t_R \sim q_{\theta}(\trm(x)|\mathcal{D})$ and computing an equally weighted average.
The predictive distribution computed as above is used to evaluate the predictive performance via RMSE.

\noindent \emph{Log-marginal Likelihood $p_{\theta}(\mathcal{Y}|\mathcal{X})$:} Lastly, we consider a Laplace approximation to the log of the marginal likelihood $p_{\theta}(\mathcal{Y}|\mathcal{X})$ as 
%
\begin{align}
    &\log g(\mathcal{Y}|\mathcal{X})\nonumber
    \\=&-\frac{1}{2}\hat{t}^{T}K_\theta(\mathcal{X})^{-1}\hat{t}+\log p(\mathcal{Y}|\trm(\mathcal{X})=\hat{t})\nonumber
    \\&-\frac{1}{2}\log |I_M+{W}^{\frac{1}{2}}K_\theta(\mathcal{X}){W}^{\frac{1}{2}}|,
\end{align} 
where $I_M$ is the identity matrix of size $M$. This is required to evaluate the weighted meta-training loss \eqref{eq:metatrainingloss} which in turn determines the Gibbs hyperposterior $q^{WFEM-GP}(\theta|\mathcal{D}_{1:N})$.

\section{Experiment Details for Regression and Classification Examples}
This section contains additional details regarding the experiments in Sec. 5. 

We instantiate the mean funtion $\mu_{\theta}(\cdot)$ and kernel function $K_{\theta}(\cdot,\cdot)$ as neural networks, where the hyperparameter $\theta$ corresponds to the weights and biases of the neural networks can be meta-learned. Both neural networks are 4 layered fully-connected neural networks with 32 neurons in each layer and tanh non-linearities. We use adaptive moment estimation (Adam) to optimize the gradient descent of updating hyperparameter $\theta$ for GP. In meta-testing phase, the performance on meta-test datasets is evaluated as detailed in Appendix~\ref{app:metatesting}.

For classification experiment, we split the mini-ImagenNet into 64 training classes, 16 validation classes and 20 testing classes, the CUB is split into 100 training classes, 50 validation classes and 50 testing classes. We evaluate the classifier via its mean prediction accuracies on test datasets. 

We have made use of the codes available at \url{https://github.com/jonasrothfuss/meta_learning_pacoh} for implementation of our algorithms.

\section{Further Experiment Results}
In Figure 5, we also investigate the impact of varying the weight parameter $\alpha$ on the performance of WFEM-GP in the regression experiment. We set $\sigma=0.1$, $\beta=0.4$, $\mu_c'-\mu_c=0.75$, $N=30$ and $M_i=5$. Tuning the weighing parameter $\alpha$ is seen to be important to optimize the accuracy. For $\beta=0.4$, the optimal performance corresponds to setting $\alpha \approx 0.2$.
\begin{figure}[H]

  \centering
  \centerline{\includegraphics[scale=0.25]{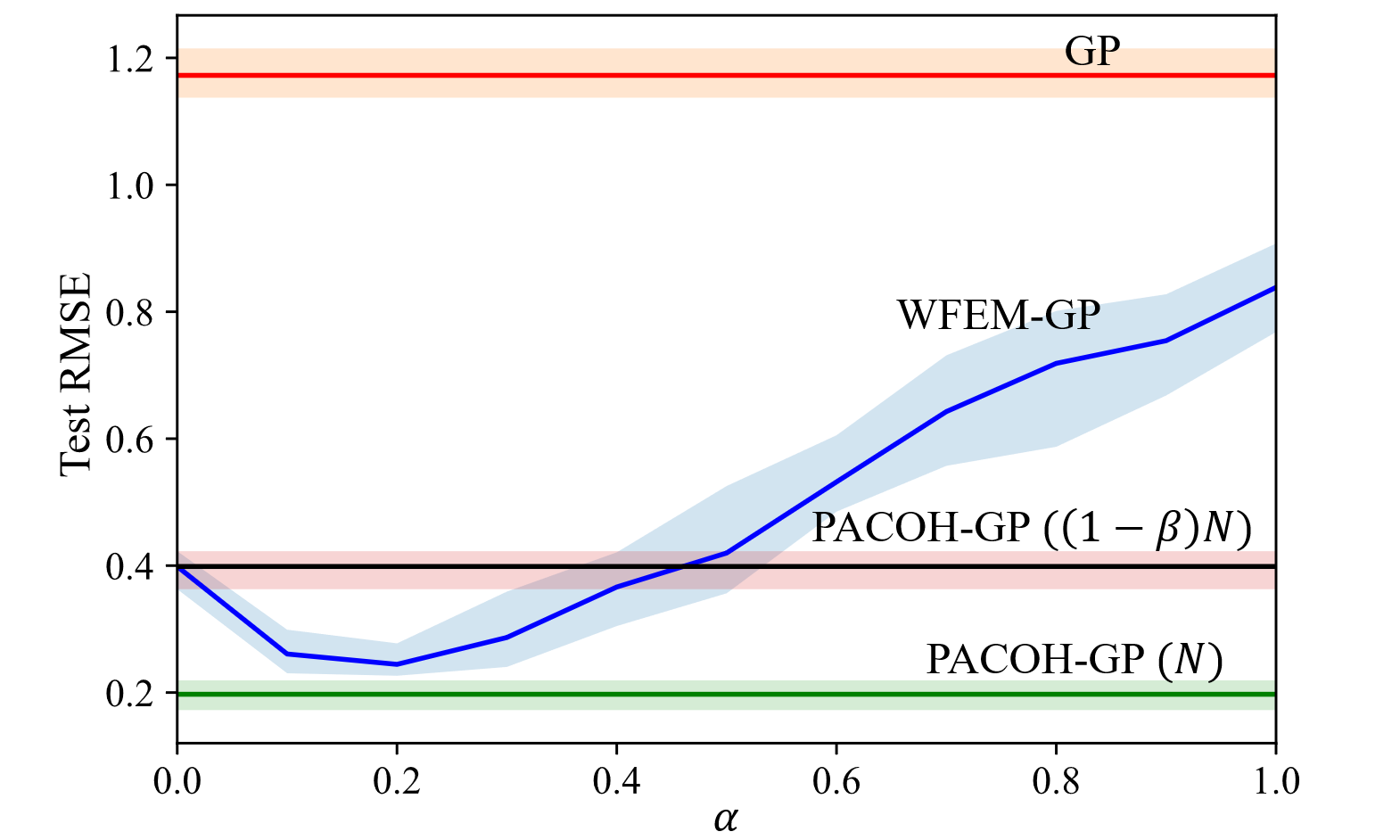}}
  \caption{Average test RMSE under four schemes -- GP, PACOH-GP with $N$ tasks and  with $(1-\beta)N$ tasks from the target environment and WFME-GP --  as a function of  $\alpha$ with fixed $\mu'_c-\mu_c=0.75$, 
  $\beta=0.4$, $\sigma=0.1$, $N=30$ and $M_i=5$.
  } 
\label{fig:res4}
\end{figure}








\end{document}